\documentclass[sigconf]{acmart}

\renewcommand\footnotetextcopyrightpermission[1]{} 
\usepackage{graphicx} 
\usepackage{subcaption} 
\usepackage{booktabs} 
\usepackage{enumitem}

\usepackage[final]{changes}  

\setcopyright{rightsretained}

\acmDOI{10.475/123_4}

\acmISBN{123-4567-24-567/08/06}

\acmConference[CACM Submission]{CACM Submission}{May 2018}{Seattle, WA USA} 
\acmYear{2018}
\copyrightyear{2018}
\acmArticle{4}
\acmPrice{15.00}

\makeatletter
\def\@copyrightspace{\relax}
\makeatother

\newcommand{\Shorten}[1]{} 
\newcommand{\Comment}[1]{} 

\newcommand{\bug}
    {\mbox{\rule{2mm}{2mm}}}

\newcommand{\etal}{\mbox{\it et al.}}

\newcommand{\eg}{\mbox{\it e.g.}}
\newcommand{\ie}{\mbox{\it i.e.}}

\newcommand{\GAM}{\mbox{GAM}}
\newcommand{\GAMsq}{\mbox{GA$^{\mbox{2}}$M}}
\newcommand{\bi}{\begin{list}{$\bullet$}{
    \setlength{\leftmargin}{1.5 em}
    \setlength{\itemsep}{0 pt}
    \setlength{\topsep}{3 pt}
    \setlength{\parsep}{3 pt}
    \setlength{\partopsep}{0 pt}
    \setlength{\labelwidth}{1 em}
    \setlength{\labelsep}{0.5 em}
    \setlength{\parskip}{0cm}  }}
\newcommand{\ei}{\end{list}}

\newcommand{\BE}{\begin{enumerate}}
\newcommand{\EE}{\end{enumerate}}

\begin{document}
\title{The Challenge of Crafting Intelligible  Intelligence}

\author{Daniel S. Weld$^{1,2}$}
\affiliation{%
  \institution{$^1$Paul G. Allen School of Computer Science \&\ Engineering\\
University of Washington\\}
  \streetaddress{185 E Stevens Way NE}
  \city{Seattle} 
  \state{Washington} 
  \postcode{98195}
}

\author{Gagan Bansal$^1$}
\affiliation{%
  \institution{$^2$Microsoft Research}
  \streetaddress{???}
  \city{Redmond} 
  \state{Washington} 
  \postcode{98195}
}








\renewcommand{\shortauthors}{}

\begin{abstract}
Since Artificial Intelligence (AI) software uses techniques like deep lookahead search and stochastic optimization of huge neural networks to fit mammoth datasets, it often results in complex behavior that is difficult for people to understand. 
Yet organizations \added{are deploying} AI algorithms in many mission-critical settings. To trust their behavior, we must make \added{AI} intelligible, either by using inherently interpretable models or by developing new methods for explaining \added{and controlling} otherwise overwhelmingly complex decisions using local approximation, vocabulary alignment, and interactive explanation.  
\added{This paper argues that intelligibility is essential, surveys recent work on building such systems, and highlights key directions for research.}
\end{abstract}

%
%


\maketitle

\section{Introduction}

Artificial Intelligence (AI) systems have reached or exceeded 
 human performance for many circumscribed tasks. As a result, they 
are increasingly deployed in mission-critical roles, such as credit scoring, predicting if a
bail candidate will commit another crime, selecting the news we read
on social networks,\Comment{Remove the news example, as it does not sound like a mission critical domain} and self-driving cars.  Unlike
other mission-critical software, extraordinarily complex AI systems are difficult to test: AI decisions are context specific and
often based on thousands or millions of factors.  Typically,
AI behaviors are generated by searching vast action spaces or learned by the opaque optimization of mammoth neural networks operating over prodigious amounts of
training data.  Almost by definition, no clear-cut method can
accomplish these AI tasks.

\begin{figure}
\centering
\includegraphics[width=3.25in]{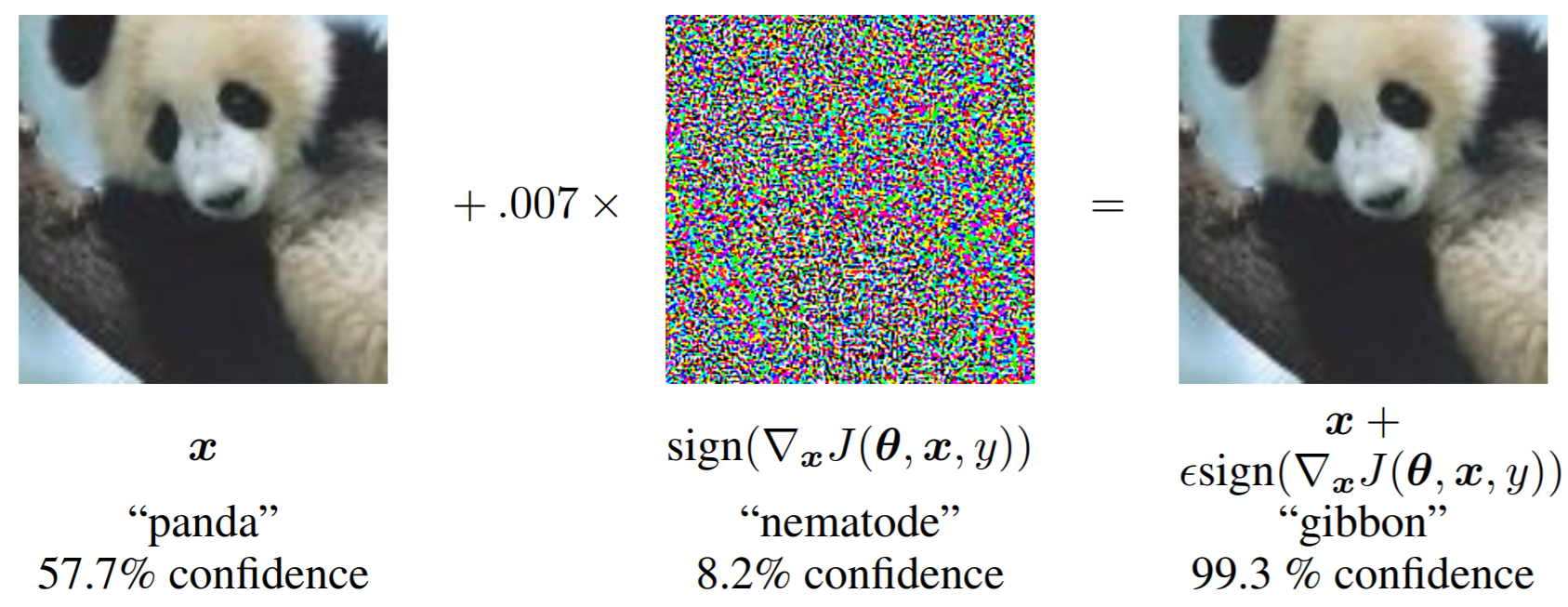}
\vspace*{-.05in}
\caption{ 
Adding an imperceptibly small vector to an image 
  changes the GoogLeNet~\cite{szegedy-cvpr15} image recognizer's classification of the image from ``panda'' to ``gibbon.'' Figure taken from Goodfellow \etal~\cite{goodfellow-iclr15}.
}
\label{f:image-rec}
\end{figure}

Unfortunately, much AI-produced behavior is {\em alien}, \ie, it can fail in unexpected ways.  This lesson is most clearly seen
in the performance of the latest deep neural network image analysis
systems. While their accuracy at object-recognition on naturally occurring pictures is extraordinary, imperceptible changes to input images can lead to
erratic predictions, as shown in
Figure~\ref{f:image-rec}. 
Why are these recognition systems so brittle, making 
different predictions for apparently identical images?  
Unintelligible behavior is not limited to machine learning; many AI programs, \added{such as automated planning algorithms,} perform search-based lookahead and  inference whose complexity exceeds human abilities to verify. \added{While some search and planning algorithms are provably complete and optimal,  intelligibility is  still important,  because the underlying primitives (\eg, search operators or action descriptions) are usually approximations~\cite{mccarthy-mi69}. One can't trust a proof that is based on (possibly) incorrect premises.} 

Despite intelligibility's apparent value,  it remains remarkably hard to
specify what makes a system ``intelligible.''
We discuss desiderata for intelligible behavior later in this article. In brief, we seek AI systems where A) it is clear what factors {\em caused} the system's action~\cite{lim-ubicomp09}, allowing the users  to predict how changes to the situation would have led to alternative behaviors, and B) permits effective control of the AI by enabling interaction.  As we will see, there is a central tension between a concise explanation and an accurate one. 

\begin{figure}
\vspace*{-0.15in}
\includegraphics[width=3.0in]{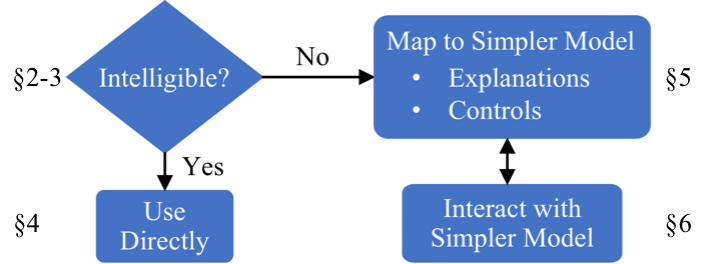}
\vspace*{-0.11in}
\caption{Approaches for crafting intelligible AI. Section numbers indicate where each aspect is discussed.
}
\vspace*{-0.1in}
\label{fig:overview}
\end{figure}

\added{As shown in Figure~\ref{fig:overview}, our survey focuses} on two high-level approaches to building intelligible AI software: 1) ensuring that the underlying reasoning or learned model is
{\em inherently interpretable}, \eg, by learning a linear model over a
small number of well-understood features, and 2) \added{if it is necessary to use an {\em inscrutable} model, such as complex neural networks or deep-lookahead search, then mapping this complex system to a simpler, {\em explanatory model} for understanding and control~\cite{lundberg-nips17}.}
Using an interpretable model provides the benefit of transparency and veracity; in theory,
a user can see exactly what the model is doing. Unfortunately, interpretable methods may not perform as
well as more complex ones, such as deep neural networks.
Conversely, the approach of mapping to an explanatory model can apply to
whichever AI technique is currently delivering the best
performance, but its explanation inherently {\em differs} from the way the AI system actually operates. \added{This yields a central conundrum:} how can a
user trust that such an explanation reflects the essence of the
underlying decision and does not conceal important details? We posit that the answer is to make the explanation system {\em interactive} so users can drill down until they are satisfied with their understanding.

The key challenge for designing intelligible AI is communicating a complex computational process to a human. This requires interdisciplinary skills, including HCI as well as AI and machine learning expertise. 
Furthermore, since the nature of explanation has long been studied by philosophy and psychology, these fields should also be
consulted.

\added{This survey highlights key approaches and challenges for building intelligible intelligence.
Section~\ref{s:doubt} characterizes intelligibility  and explains why it is important even in systems with measurably high performance. 
Section~\ref{s:gam} describes} the benefits and limitations of \GAMsq, a \added{powerful} class of interpretable ML models. \added{Then, in Section~\ref{s:explain}, we characterize methods for handling inscrutable models, discussing different strategies for mapping to a simpler, intelligible model appropriate for explanation and control. Section~\ref{s:dialog} sketches} a vision for building {\em interactive} explanation systems, \added{where the mapping changes in response to the user's needs. Section~\ref{s:planning} argues that intelligibility is important for search-based AI systems as well as for those based on machine learning and that similar solutions may be applied.}


\section{Why Intelligibility Matters}
\label{s:doubt}

\added{While it has been argued that explanations are much less important than sheer performance in AI systems,}
\added{there are many reasons why intelligibility is important.   We start by discussing technical reasons, but social factors are important as well. }

\added{{\bf AI may have the Wrong Objective:}}
In some situations, even 100\% perfect performance may be insufficient, \added{for example,} if the performance metric is flawed or incomplete due to the difficulty of specifying it explicitly. 
\added{Pundits have warned that an automated factory charged with maximizing paperclip production, could subgoal on killing humans, who are using resources that could otherwise be used in its task.
While this example may be fanciful, it illustrates that it is remarkably difficult to balance multiple attributes of a utility function.}
For example, as Lipton observed~\cite{lipton-whi16}, ``An algorithm for making hiring decisions should simultaneously optimize for productivity, ethics and legality.'' However, how does one express this trade off? 
Other examples include balancing training error while 
uncovering causality in medicine and balancing accuracy and fairness in recidivism prediction~\cite{hardt-nips16}.  
For the latter, 
a simplified objective function such as accuracy combined with historically biased training data may cause uneven performance for different groups (\eg, people of color). \added{Intelligibility empowers users to determine if an AI is right for the right reasons.}

\begin{figure*}
\centering
\includegraphics[width=6.9in]{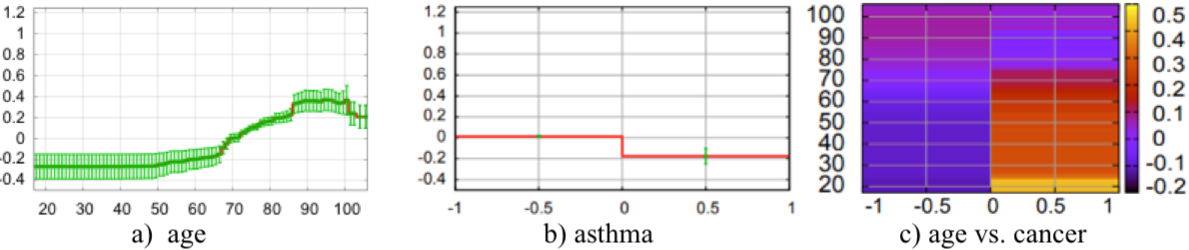}
\vspace*{-.1in}
\caption{A part of Figure 1 from \cite{caruana-kdd15} showing 3 (of 56 total) components for a \GAMsq\ model, which was trained to predict a patient's risk of dying from pneumonia. The two line graphs depict the contribution of individual features to risk: a) patient's age, and b) boolean variable asthma. The\Shorten{x-axis denotes the value of the feature and the} y-axis denotes its contribution (log odds) to  predicted risk. The heat map, c, visualizes the contribution due to pairwise interactions between age and cancer rate.
}
\label{f:gam}
\end{figure*}

\added{{\bf AI may be Using Inadequate Features:} 
Features are often correlated, and when one feature is included in a model, machine learning algorithms extract as much signal as possible from it, indirectly modeling other features that weren't included. This can lead to problematic models, as illustrated by Figure~\ref{f:gam}b (and described in the next Section), where  the ML determined that a patient's prior history of asthma (a lung disease) was {\em negatively} correlated with death by pneumonia, presumably due to correlation with (un-modeled)  variables, such as these patients receiving timely and aggressive therapy for lung problems. An intelligible model helps humans to spot these issues and correct them, \eg, by adding additional features~\cite{caruana-kdd15}.
}

{\bf Distributional Drift:} A deployed model may perform poorly \emph{in the wild}, \ie, when a difference exists between \added{the distribution which was used during training} and that encountered during deployment.
\added{Furthermore,} the deployment distribution may change over time, \added{perhaps due to feedback from the act of deployment.}  This is common in adversarial domains, such as spam detection, online ad pricing, and search engine optimization. 
\added{Intelligibility helps users determine when models are failing to generalize.}

\added{{\bf Facilitating User Control:} Many AI systems induce user preferences from their actions. For example, adaptive news feeds predict which stories are likely most interesting to a user. As robots become more common and enter the home, preference learning will become ever more common. If users understand why the AI performed an undesired action, they can better issue instructions that will lead to improved future behavior.
}

{\bf User Acceptance:} \added{Even if they don't seek to change system behavior, users have been shown to be happier with and} more likely to accept algorithmic decisions if they are accompanied by an explanation~\cite{koehler-pb91}. \added{After being told that they should have their kidney removed, it's natural for a patient to ask the doctor why --- even if they don't fully understand the answer.}

\added{{\bf Improving Human Insight:} While improved AI allows automation of tasks previously performed by humans, this is not their only use. In addition, scientists use machine learning to get insight from big data. Medicine offers several examples~\cite{caruana-kdd15}. 
Similarly, the behavior of AlphaGo~\cite{silver-nature16} has revolutionized human understanding of the game. Intelligible models greatly facilitate these processes.
}

\begin{figure}
\vspace*{-0.15in}
\includegraphics[width=2.0in]{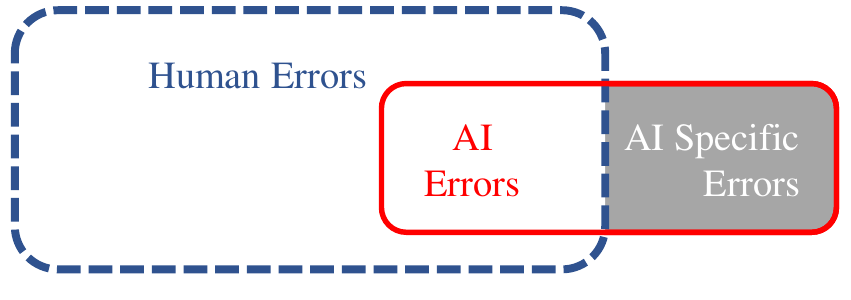}
\vspace*{-0.15in}
\caption{The dashed blue shape indicates the space of possible mistakes humans can make. The  red shape denotes the AI's mistakes; its smaller size indicates a net reduction in the  number of errors. The gray region denotes AI-specific mistakes a human would never make. 
Despite reducing the total number of errors, a deployed model may create new areas of liability (gray), necessitating explanations.
}
\vspace*{-0.18in}
\label{fig:venn}
\end{figure}

{\bf Legal Imperatives:} The European Union's GDPR legislation decrees citizens' right to an explanation,\Shorten{~\cite{goodman-eu16},} and other nations may follow. 
Furthermore, assessing legal liability is a growing area of concern;  a deployed model (\eg, self-driving cars) may introduce new areas of liability by causing accidents unexpected from a human operator, shown as ``AI-specific error'' in Figure~\ref{fig:venn}. 
Auditing such situations to assess liability, requires understanding the model's decisions.

\section{Defining Intelligibility}
\label{s:defn}

So far we have treated intelligibility informally. Indeed, 
few
 computing researchers have tried to formally define \added{what makes an AI system interpretable, transparent, or intelligible}~\cite{doshi-velez17},
but one 
suggested criterion 
is {\em human simulatability}~\cite{lipton-whi16}: can a human
user easily predict the model's output for a given input? 
By this definition, sparse linear models are more interpretable than dense or non-linear ones.

Philosophers, such as Hempel and Salmon, have long debated the nature of explanation. 
Lewis~\cite[p 217]{lewis-pp86} summarizes: ``To explain an event is to provide some
information about its causal history.'' But many causal explanations may exist.
The fact that event C causes E is best understood relative to
an imagined counterfactual scenario, where absent C, E would {\em not}
have occurred; furthermore, C should be {\em minimal}, an intuition known
to early scientists, such as William of Occam, and formalized by
Halpern and Pearl~\cite{halpern-bjps05}.

Following this logic, we suggest that a better criterion than simulatability is the ability to answer \emph{counterfactuals},   \added{aka} ``what-if'' questions. 
Specifically, we say that a model is intelligible to the degree that a human user can predict how a {\em change} to a feature, \eg, a small increase to its value, will {\em change} the model's output \added{and if they can reliably modify that response curve.}  
Note that if one can simulate the model, predicting its output, then one can predict the effect of a change, but not vice versa. 

Linear models are especially interpretable under this definition because they allow the answering of counterfactuals.
For example, consider a naive Bayes unigram model for sentiment analysis, whose objective is to predict the emotional polarity (positive or negative) of a textual passage\Shorten{~\cite{pang-ftir08}}. Even if the model were large, combining evidence from the presence of thousands of words, one could see the effect of a given word by looking at the sign and magnitude of the corresponding weight. This answers the question, ``What if the word had been omitted?''  Similarly, by comparing the weights associated with two words, one could predict the effect on the model of substituting one for the other.

\added{{\bf Ranking Intelligible Models:} Since one may have a choice of intelligible models, it is useful to consider what makes one preferable to another. 
Social science research suggests that an explanation is} best considered a social process, a conversation between explainer and explainee~\cite{hilton-psybul90,miller-explanation17}.
\added{As a result, Grice's rules for cooperative communication ~\cite{grice-maxims75} may hold for intelligible explanations. Grice's} maxim of \emph{quality} \added{says be truthful, only relating} things that are supported by  evidence. The maxim of \emph{quantity} \added{says} to give as much information as is needed, and no more.
The maxim of \emph{relation}: only say things that are relevant to the discussion.
The maxim of \emph{manner} \added{says} to avoid ambiguity, being as clear as possible. 

Miller summarizes decades of work by psychological research, noting that explanations are {\em contrastive}, \ie, of the
form ``Why P rather than Q?''  The event in question, P, is termed the
{\em fact} and Q is called the {\em foil} \cite{miller-explanation17}. Often the foil is not explicitly stated even though it is 
crucially important to the explanation process. For example, consider the question, 
``Why did you predict that the image depicts an indigo bunting?'' An explanation that points to the color blue implicitly assumes that the foil is another bird, such as a chickadee. But perhaps the questioner  wonders why the recognizer did not predict a pair of denim pants; in this case a more precise explanation might highlight the presence of wings and a beak. Clearly, an explanation targeted to the wrong foil will be unsatisfying, but the nature and sophistication of a foil can depend on the end user's expertise; hence, the ideal explanation will differ for different people~\cite{doshi-velez17}.
For example, to verify that an ML system is fair, an ethicist might generate more complex foils than a data scientist.  Most ML explanation systems have restricted their attention to elucidating the behavior of  a binary classifier, \ie, where there is only one possible foil choice. However, as we seek to explain multi-class systems, addressing this issue becomes essential.  

\added{Many systems are simply too complex to
understand without approximation. Here, the key challenge is deciding
which details to omit. After long study} psychologists determined 
that several criteria can be prioritized for
inclusion in an explanation: necessary causes (vs. sufficient ones);
intentional actions (vs. those taken without deliberation); proximal
causes (vs. distant ones); details that distinguish between fact and
foil; and abnormal features~\cite{miller-explanation17}. \Shorten{\bug should think about how to quantify these so they can be optimized}

According to Lombrozo, humans prefer explanations that are simpler
(\ie, contain fewer clauses), more general, and coherent
(\ie, consistent with what the human's prior
beliefs)~\cite{lombrozo-cogpsy07}.  In particular, she observed the
surprising result that
humans preferred simple (one clause) explanations to conjunctive
ones, even when the probability of the latter was
higher than the former~\cite{lombrozo-cogpsy07}. 
\Shorten{
Recently, \citeauthor{narayanan-arxiv18} made similar observations for explanations that used a first order logic based representation.
}
These results raise interesting questions about the purpose of explanations in
an AI system.  Is an explanation's primary purpose to convince a human
to {\em accept} the computer's conclusions (perhaps by presenting a simple, plausible, but 
unlikely explanation) or is it to {\em educate} the human
about the most likely true situation?  Tversky, Kahneman, and other psychologists have documented 
many cognitive biases that lead humans to incorrect conclusions; for example, people reason incorrectly about the {\em probability} of conjunctions, with a concrete and vivid scenario
  deemed more likely than an abstract one that strictly subsumes it~\cite{kahneman-tfas}.\Comment{Tversky \& Kahneman 1983}  Should an 
explanation system {\em exploit} human limitations or seek to {\em protect} us from them? 

Other studies raise an additional complication about how to communicate a system's uncertain predictions to human users. Koehler found that simply presenting an explanation \added{for a proposition} makes people think that it is more likely to be true~\cite{koehler-pb91}. Furthermore, 
explaining a fact in the same way as previous facts have been explained amplifies this effect~\cite{sloman-tr97}.

\section{Inherently Intelligible Models}
\label{s:gam}

 \added{Several AI systems are inherently intelligible, and we previously observed that linear models support counterfactual reasoning.}
Unfortunately, linear models have limited utility because they often result in poor accuracy. \added{More expressive choices may include simple decision trees and compact decision lists. To concretely illustrate the benefits of intelligibility, we focus on 
 \emph{Generalized additive models} (\GAM s), which are a  powerful} class of ML models that relate a set of features to \added{the} target using a linear combination of (potentially nonlinear) single-feature models called \emph{shape functions}~\cite{lou-kdd12}.
 For example, if $y$ represents the target and $\{x_1, \ldots. x_n\}$ represents the features, then a \GAM\ model takes the form $y = \beta_0 + \sum_j f_j(x_j)$, 
where the $f_i$s denote shape functions and the target $y$ is computed by summing single-feature \emph{terms}.
Popular shape functions include non-linear functions such as splines and decision trees.
\added{With linear  shape functions \GAM s  reduce to a linear models.}
\GAMsq\ models extend \GAM\ models \added{by including} 
terms for pairwise interactions between features: 
 
 \vspace*{-.11in}
 \begin{equation}
 y = \beta_0 + \sum_j f_j(x_j) + \underbrace{\sum_{i\neq j} f_{ij}(x_i, x_j)}_\text{pairwise terms}
 \end{equation}
  \vspace*{-.1in}
 
Caruana \etal\ observed that for domains containing a moderate number of \added{{\em semantic} features}, \GAMsq\ models achieve  performance that is competitive with \added{inscrutable} models, such as random forests and neural networks, while remaining intelligible~\cite{caruana-kdd15}.
Lou \etal\  observed that among  methods available for learning \GAMsq\ models, the version with bagged shallow regression tree shape functions learned via gradient boosting achieves the highest accuracy \cite{lou-kdd12}. 
 
 Both \GAM\ and \GAMsq\ are considered interpretable because the model's learned behavior can be easily understood by examining or visualizing the contribution of terms (individual or pairs of features) to the final prediction. For example, Figure~\ref{f:gam} depicts a \GAMsq\ model trained to predict a patient's risk of dying due to pneumonia, showing the contribution (log odds) to total risk for a subset of terms.
A positive contribution  increases risk, whereas a negative contribution decreases risk.\Shorten{These visualizations provide a quick and straightforward way for a domain expert to double check (and gain insight) from the model.}
For example, \added{Figure}~\ref{f:gam}(a)  shows how the patient's age affects predicted risk. While the risk is low and steady for young patients (\eg, age < 20), it increases rapidly for older patients (age > 67). Interestingly, the model shows a sudden increase at age 86; perhaps a result of less aggressive care by doctors for patients ``whose time has come.'' Even more surprising is the sudden drop for patients over 100. This might be another social effect; once a patient reaches the magic ``100'', he or she gets more aggressive care. One benefit of an interpretable model is its ability to highlight these issues, spurring deeper analysis. 

Figure \ref{f:gam}(b) illustrates another surprising aspect of the learned model; apparently, a history of asthma, a respiratory disease, {\em decreases} the patients risk of dying from pneumonia!  This finding is counter-intuitive to any physician, who recognizes that asthma, in fact, should in theory increase such risk.  When 
Caruana \etal\ checked the data, they concluded that the lower risk 
was likely due to correlated variables ---
asthma patients typically receive timely and aggressive therapy for lung issues. Therefore, although the model was highly accurate on the test set, it would likely fail, dramatically underestimating the risk to a patient with asthma who had not been previously treated for the disease.

\added{{\bf Facilitating Human Control of \GAMsq\ Models:}}
A domain expert can fix such erroneous patterns learned by the model by setting the weight of the asthma term to zero. In fact, \GAMsq s let users provide much more comprehensive feedback to the model by using a GUI to redraw a line graph for model terms \cite{caruana-kdd15}.
An alternative remedy might be to introduce a new feature to the model, representing whether the patient had been recently seen by a pulmonologist. After adding this feature, which is highly correlated with asthma,  \added{and retraining, the newly-learned} model would likely \added{reflect} that asthma \added{(by itself)} increases the risk of dying from pneumonia. 

There are two more takeaways from this anecdote. First, the {\em absence} of an important feature in the data representation can cause {\em any} AI system to learn unintuitive behavior for another, correlated feature. Second, if the learner is \added{intelligible}, then this unintuitive behavior is immediately apparent, allowing appropriate skepticism (despite high test accuracy) and easier debugging.

Recall that \GAMsq s are more expressive than simple \GAM s  because they include pairwise terms. Figure~\ref{f:gam}(c) depicts such a term for the features age and cancer. This explanation indicates that among the patients who have cancer, the younger ones are at higher risk. This may be because the younger patients who develop cancer are probably critically ill.  Again, since doctors can readily inspect these terms, they know if the learner develops unexpected conclusions.

\added{{\bf Limitations:} As described, \GAMsq\ models are restricted to binary classification, and so explanations are clearly contrastive --- there is only one choice of foil. One could extend \GAMsq\ to 
handle multiple classes by training $n$ one-vs-rest classifiers or building a hierarchy of classifiers. However, while these approaches would yield a working multi-class classifier, we don't know if they preserve model intelligibility, nor whether a user could effectively adjust such a model by editing the shape functions. }

\added{Furthermore, recall that} \GAMsq s decompose their prediction into effects of individual terms which can be visualized.
However, if users are confused about what terms mean, they will not understand the model or be able to ask meaningful ``what-if'' questions.
\added{Moreover,} if there are too many features, the model's complexity may be overwhelming. Lipton notes that the effort required to simulate some models (such as decision trees) may grow logarithmically with the number of parameters~\cite{lipton-whi16}, but for \GAMsq\ the number of visualizations to inspect \added{could} increase quadratically. Several methods \added{might} help users manage this complexity; for example, the terms \added{could} be ordered by importance; \added{however, it's not clear how to estimate importance}. Possible methods \added{include using} an ablation analysis to compute influence of terms on model performance or \added{computing} the maximum contribution of terms as seen in the training samples. Alternatively, a domain expert could group terms semantically to facilitate perusal.

However, when the number of features grows into the millions --- which occurs when dealing with classifiers over text, audio, image and video data --- existing \added{intelligible} models  do not perform nearly as well as inscrutable methods, like deep neural networks. 
Since these models combine millions of features in complex, nonlinear ways, they are beyond human capacity to simulate.

\section{Understanding Inscrutable Models}
\label{s:explain}
\label{s:map}

\added{There are two ways that an AI model may be inscrutable. It may be provided as a blackbox API,} 
such as Microsoft Cognitive Services, which uses machine learning to provide image-recognition capabilities but does not allow inspection of the underlying model. \added{Alternatively, the model may be under the user's control yet extremely complex, such as a deep,}
neural network, where a user has access to myriad learned parameters but can not reasonably interpret them. How can one best explain such models to the user?

\added{{\bf The Comprehensibility / Fidelity Trade-Off:}}
A good explanation of an event is \added{both} {\em easy to understand} and 
{\em faithful}, conveying the true cause of the event. Unfortunately, these
two criteria almost always conflict. Consider the predictions of
a deep neural network with millions of nodes: a complete and
accurate trace of the network's prediction would be far too complex
to understand, but {\em any} simplification sacrifices faithfulness. 

Finding a
satisfying explanation, therefore, requires balancing the competing goals 
of comprehensibility and fidelity. Lakkaraju \etal~\cite{lakkaraju-kdd-fatml17} suggest formulating an explicit optimization of this form and propose an approximation algorithm for generating global explanations in the form of compact sets of if-then rules. Ribeiro \etal\ describe a similar optimization algorithm that balances faithfulness and coverage in its search for summary rules~\cite{ribeiro-aaai18}. 

\added{Indeed, all methods for rendering an inscrutable model intelligible require mapping the complex model to a simpler one~\cite{lundberg-nips17}.  Several high-level approaches to mapping have been proposed.}

\added{{\bf Local Explanations:}}
\added{One} way to simplify the explanation of a learned model is to make it
relative to a single input query. Such explanations, which are termed {\em
  local}~\cite{ribeiro-kdd16} \added{or {\em instance-based}~\cite{krause-vast17}}, are akin to a doctor explaining specific 
reasons for a patient's diagnosis rather than
communicating all of her medical knowledge. \added{Contrast this approach with
the global understanding of the model that one gets with a \GAMsq\
model. Mathematically, one can see a local explanation as currying --- several variables in the model are fixed to specific values, allowing simplification. } 

Generating a local explanations is a common practice in AI systems.
For example, 
early rule-based expert systems included explanation systems that augmented a trace of the system's reasoning --- for a particular case --- with background knowledge~\cite{swartout-xplain83}. 
Recommender systems, one of the first deployed uses
of machine learning, also induced demand for explanations of their specific recommendations; the most
satisfying answers combined justifications based on the user's
previous choices, ratings of similar users, and features of the  items
being recommended~\cite{papadimitriou-dmkd12}.  

{\bf Locally-Approximate Explanations:}
In many cases, however, even a local explanation can be too complex to
understand without approximation. Here, the key challenge is deciding
which details to omit when creating the simpler explanatory model.  \added{Human preferences, discovered by psychologists and summarized in 
Section~\ref{s:defn}, should guide algorithms that construct these simplifications.
}

\begin{figure}
\centering
\includegraphics[width=2.2in]{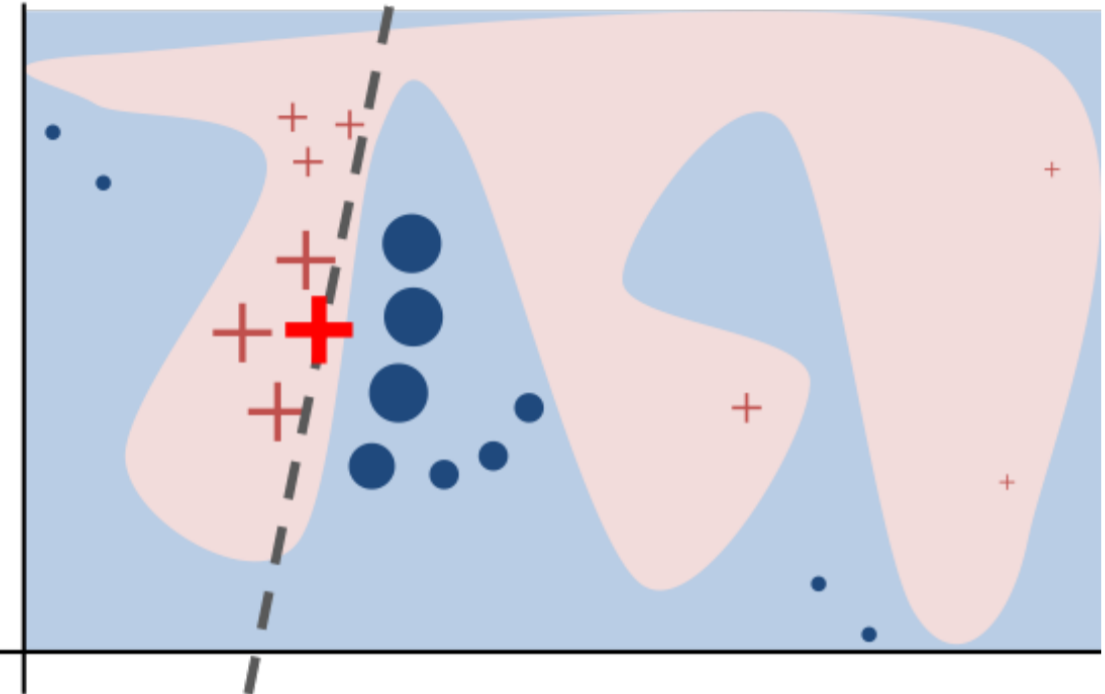}
\vspace*{-.1in}
\caption{The intuition guiding LIME's method for constructing an approximate local explanation (taken from~\cite{ribeiro-kdd16}): ``The black-box model's complex decision function, $f$,
(unknown to LIME) is represented by the blue/pink
background, which cannot be approximated well by
a linear model. The bold red cross is the instance
being explained. LIME samples instances, gets predictions
using $f$, and weighs them by the proximity
to the instance being explained (represented here
by size). The dashed line is the learned explanation
that is locally (but not globally) faithful.''
}
\label{f:lime}
\end{figure}

Ribeiro \etal's LIME system~\cite{ribeiro-kdd16} \added{is a good example of a system for generating a locally-approximate explanatory model of an arbitrary learned model, but it sidesteps part of the question of which details to omit.  Instead, LIME} requires the developer to provide two additional inputs: 1) a set of semantically meaningful features $X'$ that can be computed from the original features, and 2) an interpretable learning algorithm, such as a linear classifier (or a \GAMsq), which it uses to generate an explanation in terms of the $X'$.

The insight behind LIME is shown in Figure~\ref{f:lime}. Given an instance to explain, shown as the bolded red cross, LIME randomly generates a set of similar instances and uses the blackbox classifier, $f$, to predict their values (shown as the red crosses and blue circles). These predictions are weighted by their similarity to the input instance \added{(akin to {\em locally-weighted regression})} and used to train a new, simpler intelligible classifier, shown on the figure as the linear decision boundary, using  $X'$, the smaller set of semantic features.
The user receives the intelligible classifier as an explanation. While this {\em explanation model}~\cite{lundberg-nips17} is likely a poor global representation of  $f$, it is hopefully an accurate local approximation of the boundary in the vicinity of the instance being explained. 

Ribeiro \etal\ tested LIME on several domains. For example, they explained the predictions of a convolutional neural network image classifier by converting the pixel-level features into a smaller set of ``super-pixels;'' to do so, they ran an off-the-shelf segmentation algorithm that identified regions in the  input image and varied the color of some these regions when generating ``similar'' images. 
While LIME provides no formal guarantees about its explanations, studies showed that LIME's explanations helped users evaluate which of several classifiers best generalizes.

{\bf Choice of Explanatory Vocabulary:}
Ribeiro \etal's use of pre-segmented image regions 
to explain image classification decisions
illustrates the larger problem of determining an explanatory vocabulary. Clearly, it would not make sense to try to identify the exact pixel that led to the decision: pixels are too low level a representation and are not semantically meaningful to users. In fact, deep neural network's power comes from the very fact that their hidden layers are trained to recognize latent features in a manner that seems to perform much better than previous efforts to define such features independently. Deep networks are inscrutable exactly because we do not know what those hidden features denote.

\begin{figure}
\centering
\vspace*{-0.1in}
\includegraphics[width=3.0in]{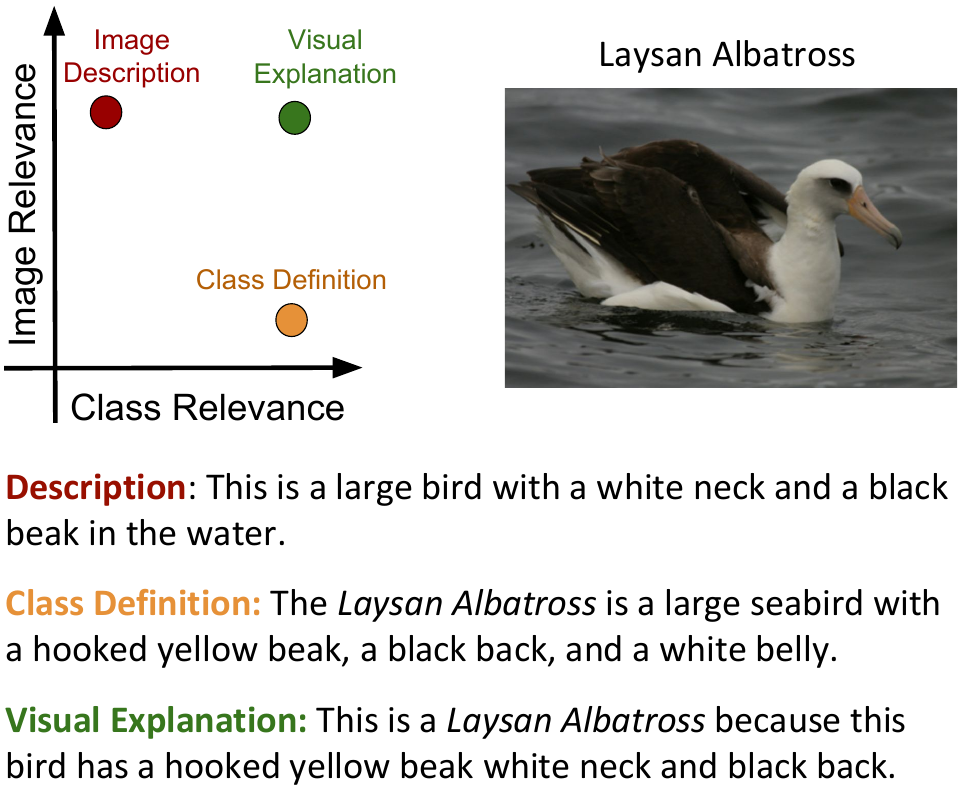}
\vspace*{-.08in}
\caption{A visual explanation taken from~\cite{hendricks-eccv16}: ``Visual explanations are both image relevant and class relevant. In contrast, image descriptions are image relevant, but not necessarily class relevant, and class definitions are class relevant but not necessarily
image relevant.''}
\label{f:visX}
\vspace*{-0.1in}
\end{figure}

To explain the behavior of such models, however, we must find some high-level abstraction over the input pixels that communicates the model's essence.  Ribeiro \etal's decision to use an off-the-shelf image-segmentation system was pragmatic. The regions it selected are easily visualized and carry some semantic value. However, regions are chosen without any regard to how the classifier makes a decision. To explain a blackbox model, where there is no possible access to the classifier's internal representation, there is likely no better option; any explanation will lack faithfulness.

However, if a user can access the classifier and tailor the explanation system to it, there are ways to choose a more meaningful vocabulary. One interesting method jointly  trains a classifier with a natural language, image captioning system~\cite{hendricks-eccv16}.  The classifier uses training data that is labeled with the objects appearing in the image; the captioning system is labeled with English sentences describing the appearance of the image. By training these systems jointly, the variables in the hidden layers may get aligned to semantically meaningful concepts, even as they are being trained to provide discriminative power.  This results in English language descriptions of images that have both high image relevance (from the captioning training data) and high class relevance (from the object recognition training data), as shown in Figure~\ref{f:visX}.

\begin{figure*}[ht]
\centering
\includegraphics[width=6in,height=1.6in]{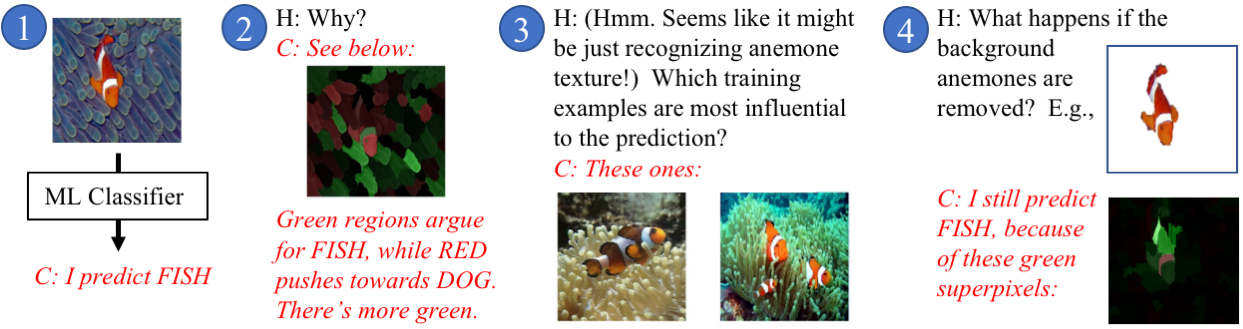}
\vspace*{-3mm}
\caption{An example of an interactive explanatory dialog for gaining insight into a DOG/FISH image classifier.  (For illustration, the questions and answers are shown in English language text, but our use of a `dialog' is for illustration only. An interactive GUI, \added{e.g., building on the ideas of Krause et al.~\cite{krause-vast17}, would likely}  be a better realization.)}
\label{f:dialog}
\vspace*{-2mm}
\end{figure*}

While this method works well for many examples, some explanations  include details that are not actually present in the image; newer approaches, such as phrase-critic methods, may create even better descriptions~\cite{hendricks-nips-iml17}.
Another approach might determine if there are hidden layers in the learned classifier that learn concepts corresponding to something meaningful. For example, \citeauthor{zeiler-eccv14} observed that certain layers may function as edge or pattern detectors~\cite{zeiler-eccv14}. Whenever a user can identify the presence of such layers, then it may be preferable to use them in the explanation. Bau \etal\ describe an automatic mechanism for matching CNN representations with semantically meaningful concepts using a large, labeled corpus of objects, parts, and texture; furthermore, using this alignment, their method quantitatively scores CNN interpretability, potentially suggesting a way to optimize for intelligible models.  

However, many obstacles remain. As one example, it is not clear that there {\em are} satisfying ways to describe important, discriminative features, which are often intangible, \eg, textures. An intelligible explanation may need to define new terms or combine language with other modalities, like patches of \added{an} image. 
\added{Another challenge is inducing first-order, relational descriptions, which would enable descriptions such as ``a spider because it has eight legs'' and ``full because all seats are occupied.''
While quantified and relational abstractions are very natural for people, progress in statistical-relational learning has been slow and there are many open questions for neuro-symbolic learning~\cite{besold-neural-symbolic18}.
}

\Comment{shouldn't one think of borrowing or augmenting GOFAI principles for ML? (Eg cf. statistical relational learning, or unifying symbolic and learning methods.}


{\bf Facilitating User Control with Explanatory Models:}
Generating an explanation by mapping an inscrutable model into a simpler, explanatory model is only half of the battle. In addition to answering counterfactuals about the original model, we would ideally be able to   map any control actions the user takes in the explanatory model back as adjustments to the original, inscrutable model.  
For example, Section~\ref{s:gam} illustrated how a user could directly edit a \GAMsq's shape curve (Figure~\ref{f:gam}(b)) to change the model's response to asthma. Is there a way to interpret such an action, made to an intelligible explanatory model, as a modification to the original, inscrutable model?   
It seems unlikely that we'll discover a general method  to do this for arbitrary source models, since the abstraction mapping is not invertible in general. However, there are likely methods for mapping backwards to specific classes of source models or for specific types of feature-transform mappings. This is an important area for future study.

\section{Towards Interactive Explanation} 
\label{s:dialog}

The optimal choice of explanation depends on the audience. 
 Just as a human teacher would explain physics differently to students
    who know or do not yet know calculus, the technical sophistication and
    background knowledge of the recipient affects the suitability of a
    machine-generated explanation. 
\added{Furthermore, the concerns of a house seeker whose mortgage application was denied due to a FICO score differ from those of a developer or data scientist debugging the system.} 
\added{Therefore, an ideal explainer should model the} user's background over the course of many interactions.
 
\added{The HCI community has long studied mental models~\cite{norman14}, and many intelligent tutoring systems (ITSs) build explicit models of students' knowledge and misconceptions~\cite{anderson-science85}.  However, the framework for these models are typically hand-engineered for each subject domain, so it may be difficult to adapt ITS approaches to a system that aims to explain an arbitrary black-box learner.}

\added{Even with an accurate user model, it is likely that an explanation will not answer all of a user's concerns, because the human may have follow-up questions.
We conclude that an explanation system should be {\em interactive}, supporting such questions from and actions by the user. 
This matches results from psychology literature, summarized in Section~\ref{s:doubt}, and highlights Grice's maxims, especially those pertaining to quantity and relation.  It also builds on Lim and Dey's work in ubiquitous computing, which investigated the kinds of questions users wished to ask about complex, context-aware applications~\cite{lim-ubicomp09}.}  We envision an interactive explanation system that supports many different follow-up and drill-down action after presenting a user with an initial explanation:

\bi
\item {\em Redirecting the answer by changing the foil}: ``Sure, but why didn't you predict class C?''

\item {\em Asking for more detail} (\ie, a more complex explanatory model), perhaps while restricting the explanation to a subregion of feature space: ``I'm only concerned about women over age 50...'' 

\item {\em Asking for a decision's rationale}: ``What made you believe this?'' To which the system might respond by displaying the labeled {\em training examples} that were most influential in reaching that decision, \eg, ones identified by influence functions~\cite{koh-icml17} or nearest neighbor methods.

\item {\em Query the model's sensitivity} by asking what minimal perturbations to certain features would lead to a different output.

\item {\em Changing the vocabulary} by adding (or removing) a feature in the explanatory model, either from a predefined set, by using methods from machine teaching\Shorten{~\cite{simard-machine-teach17}}\added{, or with concept activation vectors~\cite{kim-tcav-arxiv17}.}
\Shorten{For example, instead of super-pixels computed using an automated algorithm \cite{Ribeiro2016}, a user may prefer a representation grounded in an intermediate representation used by the system, or an existing external knowledge base \cite{Chen2015MicrosoftCC}, or multiple modalities \cite{Hendricks2016}. 
A dialog-based system should also permit human-supplied representations by supporting the interactive definition of new terms and then presentation of new explanations in terms of an augmented vocabulary.}

\item {\em Perturbing the input example} to see the effect on both prediction and explanation. In addition to aiding understanding of the model \added{(directly testing a counterfactual), this action enables an affected user who} wants to contest the initial prediction: ``But officer, one of those prior DUIs was overturned...?''

\item \added{ {\em Adjusting the model:} Based on new understanding, the user may wish to correct the model. Here, we expect to build on tools for interactive machine learning~\cite{amershi-aim14} 
and explanatory debugging~\cite{kulesza-iui15,krause-vast17}, which have explored interactions for adding  new training examples, correcting  erroneous labels in existing data, specifying new features, and modifying shape functions. As mentioned in the previous section, it may be challenging to map user adjustments, that are made in reference to an explanatory model, back into the original, inscrutable model.
}

\ei

To make these ideas concrete, Figure~\ref{f:dialog} presents a possible dialog as a user tries to understand the robustness of a deep neural dog/fish classifier built atop Inception v3~\cite{szegedy-cvpr15}. As the figure shows: 1) The computer correctly predicts that the image depicts a fish. 2) The user requests an explanation, which is provided using LIME~\cite{ribeiro-kdd16}. 3) The user, concerned that the classifier is paying more attention to the background than to the fish itself, asks to see the training data that influenced the classifier; the nearest neighbors are computed using influence functions~\cite{koh-icml17}. While there are anemones in those images, it also seems that the system is recognizing a clownfish. 4) To gain confidence, the user edits the input image to remove the background, resubmits it to the classifier and checks the explanation.

\section{Explaining Combinatorial Search}
\label{s:planning}

Most of the preceding discussion has focused on intelligible {\em machine learning}, which is just one type of artificial intelligence. However, the same issues also confront systems based on {\em deep-lookahead search}. 
While many planning algorithms have strong theoretical properties, such as soundness, they search over action {\em models} that include their own assumptions. Furthermore, goal specifications are likewise incomplete~\cite{mccarthy-mi69}. If these unspoken assumptions are incorrect, then a formally correct plan may still be disastrous.

Consider a planning algorithm that has generated a sequence of actions for a remote, mobile robot. If the plan is short with a moderate number of actions, then the problem may be inherently intelligible, and a human could easily spot a problem.
However, larger search spaces could be cognitively overwhelming.  In these cases, local explanations offer a simplification technique that is helpful, just as it was when explaining machine learning. The vocabulary issue is likewise crucial: how does one succinctly and abstractly summarize a complete search subtree?  Depending on the choice of explanatory foil, different answers are appropriate~\cite{fox-ijcai-xai17}.
\citeauthor{sreedharan-draft18} describe an algorithm for generating the minimal explanation that patches a user's partial understanding of a domain~\cite{sreedharan-draft18}. \added{Work on mixed-initiative planning~\cite{ferguson-aaai98} has demonstrated the importance of supporting interactive dialog with a planning system.}
Since many AI systems, \eg, AlphaGo~\cite{silver-nature16}, combine  deep search {\em and} machine learning, 
additional challenges will result from the need to explain interactions between combinatorics and learned models.

\section{Final Thoughts}

\added{In order to trust deployed AI systems, we must not only improve their robustness~\cite{dietterich-aimag17}, but  also develop ways
to make their reasoning intelligible. Intelligibility will help us spot AI that makes mistakes due to distributional drift or incomplete representations of goals and features.
Intelligibility will also facilitate control by humans in increasingly common collaborative human/AI teams. Furthermore, intelligibility will help humans learn from AI.  Finally, there are legal reasons to want intelligible AI, including the European GDPR and a growing need to assign liability when AI errs.}

Depending on the complexity of the models involved, two approaches to enhancing understanding may be appropriate: 1) using an inherently interpretable model, or 2) adopting an inscrutably complex model and generating  post hoc explanations by \added{mapping it to a simpler, explanatory model through a combination of currying and local approximation.} When learning a model over a medium number of human-interpretable features, one may confidently balance performance and intelligibility with approaches like \GAMsq s. However, for problems with thousands or millions of features, performance requirements likely force the adoption of inscrutable methods, such as deep neural networks or boosted decision trees. In these situations, post-hoc explanations may be the only way to facilitate human understanding.

Research on explanation algorithms is developing rapidly, with work on both local (instance-specific) explanations and global approximations to the learned model. A key challenge for all these approaches is \added{the} construction of an explanation vocabulary, essentially a set of features used in the approximate explanation model. \added{Different explanatory models may be appropriate for different choices of explanatory foil, an aspect deserving more attention from systems builders. 
While many intelligible models can be directly edited by a user, more research is needed to determine how best to map such actions back to modify an underlying inscrutable model.}
Results from psychology show that explanation is a social process\added{, best thought of as a conversation.} As a result, we advocate increased work on {\em interactive} explanation systems that support a wide range of follow-up actions. To spur rapid progress in this important field, we hope to see collaboration between researchers in multiple disciplines.

\Comment{

* GAMsq interpretability in multi-class case

issue: only locally-approximate, choosing vocabulary (super-pixels, multi-modal etc)

future directions: none of the approaches consider audience background- explanation needs of data scientist and end user differ.

* evaluation will  require user studies - Challenges are both HCI and core AI in nature

* dialog

* vocabulary

* control - mapping adjustments back to original model

So can we ever trust these AI systems?
Well,   we trust other humans even though they aren't transparent.

Indeed, Kahneman TFS notes that the processes by which humans make
decisions and explain them are often different.

directions for the future

* vocabulary

* dialog
}

{\bf Acknowledgements:}
 \added{We thank  E. Adar, S. Ameshi, R. Calo, R. Caruana, M. Chickering, O. Etzioni, J. Heer, E. Horvitz, T. Hwang, 
  R. Kambhamapti, E. Kamar,  S. Kaplan, B. Kim, P. Simard, 
 Mausam,  C. Meek, M. Michelson, S. Minton, B. Nushi, G. Ramos,
 M. Ribeiro, M. Richardson, P. Simard, J. Suh, J. Teevan,  T. Wu, and the anonymous reviewers for helpful conversations and comments. This work was supported in part by the Future of Life Institute grant 2015-144577 (5388)
with additional support from NSF grant IIS-1420667, ONR grant N00014-15-1-2774, and the WRF/Cable Professorship.}



\bibliographystyle{ACM-Reference-Format}
\bibliography{xai}

\end{document}